\documentclass[conference]{IEEEtran}
\IEEEoverridecommandlockouts
\usepackage[utf8]{inputenc}
\usepackage{graphicx}
\usepackage{amsmath, amssymb, amsfonts}
\usepackage{booktabs}
\usepackage{tabularx}
\usepackage{float}
\usepackage{algorithm}
\usepackage{algpseudocode}
\usepackage{multirow}
\usepackage{caption}
\usepackage{subcaption}
\usepackage{enumitem}
\usepackage{cite}
\usepackage{xcolor}
\usepackage{tikz}
\usetikzlibrary{positioning}
\usepackage{hyperref}
\usepackage{comment}
\usepackage{tcolorbox}
\usepackage{todonotes}
\hypersetup{colorlinks=true, linkcolor=blue, citecolor=blue, urlcolor=blue}

\title{Learning the Pareto Space of Multi-Objective Autonomous Driving: A Modular, Data-Driven Approach}

\author{\IEEEauthorblockN{Mohammad Elayan}
\IEEEauthorblockA{\textit{Dept. of Civil \& Environmental Engineering} \\
\textit{University of Nebraska-Lincoln}\\
melayan2@nebraska.edu}
\and
\IEEEauthorblockN{Wissam Kontar$^*$}
\IEEEauthorblockA{\textit{Dept. of Civil \& Environmental Engineering} \\
\textit{University of Nebraska-Lincoln}\\
wkontar2@nebraska.edu}
}

\begin{document}

\maketitle

\begin{abstract} 
Balancing safety, efficiency, and interaction is fundamental to designing autonomous driving agents and to understanding autonomous vehicle (AV) behavior in real-world operation. This study introduces an empirical learning framework that derives these trade-offs directly from naturalistic trajectory data. A unified objective space represents each AV timestep through composite scores of safety, efficiency, and interaction. Pareto dominance is applied to identify non-dominated states, forming an empirical frontier that defines the attainable region of balanced performance.

The proposed framework was demonstrated using the Third Generation Simulation (TGSIM) datasets from Foggy Bottom and I-395. Results showed that only 0.23\% of AV driving instances were Pareto-optimal, underscoring the rarity of simultaneous optimization across objectives. Pareto-optimal states showed notably higher mean scores for safety, efficiency, and interaction compared to non-optimal cases, with interaction showing the greatest potential for improvement.

This minimally invasive and modular framework, which requires only kinematic and positional data, can be directly applied beyond the scope of this study to derive and visualize multi-objective learning surfaces. 

\emph{Full reproducibility is supported via our open-source codebase on \href{https://github.com/wissamkontar/Learning-the-Pareto-Space-of-Multi-Objective-Autonomous-Driving}{GitHub}.}

\end{abstract}


\section{Introduction}

Automated driving systems are steadily transitioning from experimental deployments to complex, mixed-traffic environments. Yet, the behavioral footprint of these vehicles remains only partially understood. Global research on autonomous driving is increasingly moving toward the development of driving agents that learn and adapt through multi-objective reasoning. These consensus-driven agents aim to balance performance dimensions rather than optimize a single goal, which marks a shift toward socially aware, context-adaptive autonomy. Achieving such balance is central to creating driving agents that integrate smoothly within human-dominated traffic systems.

Human drivers continuously navigate competing objectives, such as maintaining safety margins without impeding flow or signaling intent through subtle motion cues. AVs, by contrast, rely on a mix of rule-based and learned control policies that integrate perception, planning, and decision-making. While these systems achieve consistent and safe operation, their design constraints and conservative tuning can limit efficiency or interaction fluidity. As AV deployment expands, understanding this multi-objective balance is essential for defining the attainable space of context-sensitive, high-performance operation, rather than raking AVs by a single score.

This study introduces an empirical learning framework that derives multi-objective trade-offs directly from naturalistic trajectory data. Our method was tested on the Third Generation Simulation (TGSIM) data, which capture diverse urban driving contexts and mixed traffic interactions. A unified objective space across safety, efficiency, and interaction represents each AV driving instance as a point within a behavioral landscape, from which the Pareto-optimal frontier is learned using Gaussian Process Regression (GPR). The resulting surface captures the continuous structure of optimal trade-offs while suppressing noise and segmentation artifacts.

\begin{tcolorbox}[colback=green!6!white,colframe=green!40!black]
Learning the Pareto space means identifying behavioral states where improving one objective (safety, efficiency, or interaction) necessarily reduces another. It represents a continuous surface of optimal trade-offs derived from real-world data, forming the basis for learning balanced and adaptive autonomous driving.
\end{tcolorbox}

\begin{figure}[ht!]
\centering
\includegraphics[width=0.98\linewidth]{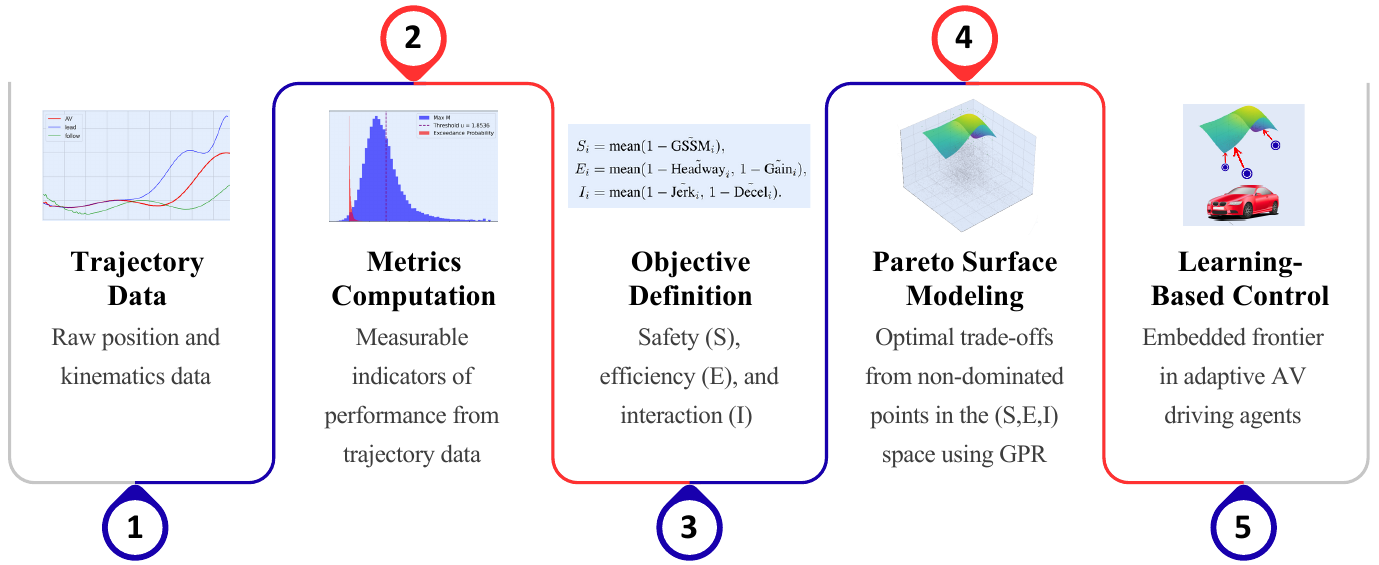}
\caption{Conceptual overview of the overarching framework linking empirical data to learning-based control.}
\label{fig:flow_learning}
\end{figure}

The framework situates empirical behavior within the broader pursuit of adaptive, learning-based driving agents. Beyond evaluation, it provides a transferable foundation for multi-objective learning and control (Figure~\ref{fig:flow_learning}. Its modular structure, built on basic positional and kinematic inputs and requiring no specialized instrumentation, can be applied to any trajectory dataset or metric formulation. It connects empirical observation with prescriptive modeling, bridging naturalistic data and the design of consensus-driven autonomous systems.

\section{Literature Review}

AVs display behavior that diverges systematically from that of human-driven vehicles (HDVs). Empirical studies consistently show that AVs maintain wider spatial buffers and higher values of surrogate safety indicators such as time-to-collision (TTC) and post-encroachment time (PET), signaling a cautious, safety-prioritized control logic \cite{Saeed_Rahmani_2024, Zou2023-dn, Hammami2024-ul}. Yet, traditional surrogate safety measures (SSMs) like TTC and PET, which were developed around human perceptual and reaction constraints, are not directly transferable to mixed-traffic environments. These metrics overlook the low-latency perception–response loops and rule-based decision layers characteristic of AV systems \cite{Tafidis2023-cn, Zhang2022-jo}. Consequently, short TTC values may not necessarily imply risk in automated contexts, while PET thresholds can mischaracterize the outcome of negotiated or non-conflict interactions \cite{Xhoxhi2024-zs}.  

Beyond safety, AV operation influences broader traffic dynamics. Controlled experiments and field data reveal that AV-led platoons can suppress oscillations and improve local stability \cite{Luo2024-bk, Hu2021-kq}. However, under heterogeneous conditions, cautious accelerations and large headways may reduce overall throughput \cite{Das2024-tr, kontar2021multi}. Recent studies also highlight AVs’ lack of visual and social cues compared to HDVs, which can affect pedestrian trust and increase gap acceptance times \cite{Izquierdo2024-yu, Lau2024-ex}.

Recent research trends move toward such integrated evaluation. Composite indicators like SAFR-AV and risk factor scores attempt to encapsulate safety and interaction jointly \cite{Pathrudkar2023-xo, Xhoxhi2024-zs}, while Pareto-based modeling in simulation studies has illuminated recurring trade-offs among comfort, efficiency, and safety \cite{9916133, 9966123, li2023pareto}. Parallel algorithmic literature frames these trade-offs explicitly as multi-objective optimization or reinforcement learning (RL) tasks, enabling agents to negotiate among competing goals via constrained or Pareto-based reward formulations \cite{10580383, yang2024}. Advances in multi-agent and consensus-oriented RL further demonstrate that cooperative adaptation among AVs can enhance both safety and stability at the network scale \cite{Vinitsky2023Optimizing, kazemkhani2025}.  

Despite advances in AV research, empirical understanding of how these systems manage multi-objective trade-offs in real traffic remains limited. Most prior studies rely on simulations or theoretical control models. This study constructs an \emph{empirical Pareto frontier} directly from observed trajectories, evaluating AV behavior across safety (via GSSM), interaction, and efficiency. The framework quantifies how real-world driving aligns with optimal balance and provides a basis for benchmarking behavioral consensus and informing next-generation AV decision design.

\section{Dataset Description}
This study uses two high-resolution trajectory datasets from the \textit{Third Generation Simulation (TGSIM)} project—Foggy Bottom and I-395, collected by the U.S. Department of Transportation’s Intelligent Transportation Systems Joint Program Office and released in January 2025 \cite{fhwa2025tgsim} \cite{fhwa2025tgsim_i395}.

The \textbf{Foggy Bottom} dataset captures trajectories at four urban intersections in Washington, D.C. It includes pedestrians, cyclists, scooters, passenger cars, trucks, buses, and \textbf{SAE Level 3 AVs}. AVs completed four consecutive left turns, providing a consistent basis for assessing performance in dense traffic. The \textbf{I-395} dataset covers a 0.5 km freeway segment in Washington, D.C. It includes a major on-ramp and off-ramp, capturing frequent merges and lane-keeping by passenger cars, trucks, buses, and 21 \textbf{SAE Level 2 AVs} operating exclusively in one lane. Each record includes timestamped positions, lane IDs, dimensions, and kinematics. 

For both datasets, pixel coordinates were transformed to a meter-scale frame with the origin at the lower-left reference corner. Kinematic variables were smoothed with Gaussian filters to preserve transient events such as braking and turning. All behavioral metrics were derived from the filtered \textit{x–y} components for consistency. The full preprocessing workflow is publicly available at our \emph{\textcolor{blue}{\href{https://github.com/wissamkontar/Learning-the-Pareto-Space-of-Multi-Objective-Autonomous-Driving}{GitHub
 repository}}}.

\section{Methodology}

In this section, we present the methodological framework used to analyze AV behavior and construct the Pareto frontier. 

\subsection{Identifying Detected Agents}
Tesla’s camera-based \textit{Tesla Vision} detection zones were adopted as a spatial reference to identify interactions between AVs and surrounding agents. These zones, while not identical to the TGSIM AV sensors, offer a standardized and empirically supported basis for proximity and angular coverage.

At each time step, Euclidean distances and relative angles between the AV and nearby agents were computed. Agents were then assigned to the corresponding detection zones based on range and field-of-view thresholds.


\subsection{Affine Spacing-Based Interaction Identification}

At each timestep, potential followers of the ego AV were identified using the affine spacing policy, which defines dynamic coupling between vehicles \cite{Michael_Shaham_2024}. A vehicle \( j \) was considered an effective leader or follower if its longitudinal spacing relative to the ego vehicle \( i \) satisfied:

\begin{equation}
    |d_{\text{actual}} - (d_0 + h v_i)| \leq \epsilon,
\end{equation}

where \( d_{\text{actual}} \) is the Euclidean distance, \( d_0 = 4\,\text{m} \) is the standstill distance, \( h = 2.0\,\text{s} \) the desired headway, and \( \epsilon = 5\,\text{m} \) a noise tolerance \cite{Ahn2007,Nowell2000}. Only vehicles detected in the rear zone and satisfying this condition were retained for string stability calculations. 

\subsection{Metrics for Behavioral Consensus}

To evaluate AV behavior across contexts, we derived a suite of surrogate safety, interaction, and efficiency metrics that define the Pareto surface. This translates the objectives of autonomous driving into measurable quantities, enabling systematic analysis of safety, efficiency, and interaction. Here, we focus on the selected metrics within our framework. A detailed examination of all candidate metrics considered prior to this final set is provided in our earlier work \cite{ElayanKontar2025Pareto}.

\subsubsection{Generalized Surrogate Safety Measure (GSSM)}

We developed a variant of the GSSM \cite{jiao2025}, adapted to the structure of the TGSIM datasets. GSSM quantifies continuous, context-aware collision risk from multi-directional interactions, extending beyond one-dimensional metrics such as TTC or PET.  

For each AV–agent pair $(i,j)$, relative distance $s_{ij}$, angle $\rho_{ij}$, and relative speed $|\mathbf v_{ij}|$ were calculated in the ego vehicle’s coordinate frame. Only agents within the Tesla Vision detection zones were retained. Each interaction was represented by a feature vector $X$ capturing relative motion, geometry, agent type, and roadway context. Conditional spacing $p(s|X)$ was modeled as lognormal, with $(\mu,\sigma)$ estimated via a neural network to capture nonlinear, context-dependent effects and corrected for heteroskedasticity via a residual term. The model achieved a mean absolute error of 6.4\,m.   

The survival probability $P(S>s^*|X)$, derived from the learned conditional spacing distribution, defines a context-conditioned surrogate risk score $M(s^*,X)$ 

\begin{equation}
M(s^*,X)=\log_{10}\!\left(\frac{\ln 0.5}{\ln P(S>s^*|X)}\right),
\end{equation}
where $M>0$ indicates elevated risk.  

To characterize rare, high-risk events, the upper tail of $M$ was modeled using \emph{Extreme Value Theory} with a Generalized Pareto Distribution (GPD). Following \cite{Scarrott_MacDonald_2012}, stability analysis identified the 97th percentile ($u=1.85$) as the optimal threshold for valid tail behavior. For exceedances $M>u$, the probability of more extreme risk is  

\begin{equation}
P(M' > M)=\left(1+\frac{\xi(M-u)}{\beta}\right)^{-1/\xi}
\end{equation}

\subsubsection{Headway Estimation}

Headway was defined at each timestep as the Euclidean distance between the ego AV and its nearest valid leader within the forward vision zone and along the same or adjacent route segment:

\begin{equation}
H_i(t) = d_{\text{actual}}(i,L_i)
\end{equation}

Leader–follower pairs were validated following the Ultra-AV criteria \cite{Zhou2024}, ensuring motion continuity ($v_f>0.1$\,m/s), perceptual proximity ($d<120$\,m), realistic accelerations ($|a_f|\!<\!5$\,m/s$^2$), and consistent relative motion.

\subsubsection{String Stability Gain}

For each AV, the nearest valid follower was identified to assess disturbance propagation. The follower’s reaction delay ($\tau$) was estimated by cross-correlating acceleration histories within 5\,s windows and refined using a context-sensitive Generalized Additive Model:

\begin{equation}
\tau \sim s(v_l)+s(v_f)+s(d)+s(a_l)+f(\text{lane})+f(\text{type}_f)
\end{equation}

(\textit{Pseudo}-$R^2$=0.44). String stability gain was  computed as:

\begin{equation}
G_i(t)=\frac{a_{F_i}(t-\tau_i)}{a_i(t)},
\end{equation}

where $G_i(t)\!\le\!1$ denotes stable behavior. This context-aware method avoids misalignment in fixed-delay approaches.

\subsubsection{Jerk and Deceleration Metrics}

Maneuver smoothness was quantified by the jerk magnitude, which is calculated as the time derivative of acceleration (${\Delta t = 0.1s}$).

Smoothed longitudinal and lateral accelerations were used to compute jerk; values exceeding $2.5$\,m/s$^3$ indicated jerky behavior \cite{Genser_2021, FENG2017125}.  
Deceleration events sustained for at least three frames and exceeding $2.0$\,m/s$^2$ were flagged as uncomfortable or pressured responses \cite{Moon01082008, bae2022selfdrivinglikehumandriver}.

\subsection{Empirical Pareto Optimality Analysis}

To capture multi-objective trade-offs beyond threshold-based consensus, we constructed an empirical Pareto frontier across safety, efficiency, and interaction. Each event was represented by a 3D objective vector:

\begin{equation}
    \mathbf{x}_i = (S_i, E_i, I_i),
\end{equation}

where \(S_i\), \(E_i\), and \(I_i\) are composite scores derived from normalized safety, efficiency, and interaction metrics. All metrics were min–max normalized within context:

\begin{equation}
    \tilde{m}_i = \frac{m_i - \min(m)}{\max(m) - \min(m)},
\end{equation}

and inverted as \(1 - \tilde{m}_i\) when lower values indicated better performance. Objective scores were computed as the mean of relevant normalized metrics:

\begin{equation}
\label{eq:pareto_scores}
\begin{aligned}
    S_i &= \text{mean}(1 - \tilde{\text{GSSM}}_i), \\
    E_i &= \text{mean}(1 - \tilde{\text{Headway}}_i,\, 1 - \tilde{\text{Gain}}_i), \\
    I_i &= \text{mean}(1 - \tilde{\text{Jerk}}_i,\, 1 - \tilde{\text{Decel}}_i)
\end{aligned}
\end{equation}

Missing metrics, occasionally caused by lack of leader and/or follower, were imputed using distance-weighted k-nearest neighbors (KNN) within the same dataset group (Foggy Bottom or I-395).

Let \(\mathcal{X} = \{\mathbf{x}_1, \dots, \mathbf{x}_n\}\) denote the set of all $n$ observed interaction events. \(\mathbf{x}_a\) dominates \(\mathbf{x}_b\) (\(\mathbf{x}_a \prec \mathbf{x}_b\)) if:

\begin{equation}
\label{eq:pareto_rule}
(S_a,E_a,I_a) \ge (S_b,E_b,I_b)\; \text{and}\; (S_a,E_a,I_a) \ne (S_b,E_b,I_b) 
\end{equation}

The Pareto-optimal set is:

\begin{equation}
\label{eq:pareto_set}
    \mathcal{P} = \{\mathbf{x}_i \in \mathcal{X}\;|\;\nexists\, \mathbf{x}_j \in \mathcal{X}:\mathbf{x}_j \prec \mathbf{x}_i\}
\end{equation}

Equation (\ref{eq:pareto_rule}) defines dominance, whereby an event dominates another if it is at least as good in all objectives and strictly better in at least one. Equation (\ref{eq:pareto_set}) identifies the subset of events not dominated by any other, forming the Pareto-optimal set. A GPR was then fitted to the Pareto-optimal points to form a smooth, continuous frontier. This surface approximates the empirical trade-off structure among safety, efficiency, and interaction, revealing how real-world AV behavior distributes across optimal balance regions.

\begin{algorithm}[!ht]
\caption{Empirical Pareto Frontier Construction}
\label{alg:pareto}
\begin{algorithmic}[1]
\Statex \textbf{Input:} Metric dataset $\mathcal{D}$
\Statex \textbf{Output:} Pareto-optimal set $\mathcal{P}$ and Pareto surface
\vspace{0.5em}

\Statex \textit{\textbf{Data Preparation}}
\State Import trajectory data
\State Derive metrics from trajectory data
\State Merge metrics by $(\text{av\_id}, \text{time})$ to construct $\mathcal{D}$
\State Filter invalid or extreme metric values in $\mathcal{D}$
\State Normalize all metrics to $[0,1]$ and invert (lower is better)

\vspace{0.5em}
\Statex \textit{\textbf{Objective Construction and Pareto Identification}}
\State Compute composite objectives (Equation~\ref{eq:pareto_scores})
\State Impute missing objectives using contextual similarity
\State Identify Pareto-optimal set $\mathcal{P}$ (Equations~\ref{eq:pareto_rule} and~\ref{eq:pareto_set})

\vspace{0.5em}
\Statex \textit{\textbf{Frontier Modeling}}
\State Fit a smooth frontier surface to $\mathcal{P}$ using GPR
\State \textbf{Return:} $\mathcal{P}$ and fitted frontier
\end{algorithmic}
\end{algorithm}

\section{Results}
\subsection{Behavioral Assessment Results}

\subsubsection{GSSM}
\begin{figure}[ht!]
\centering
\includegraphics[width=0.56\linewidth]{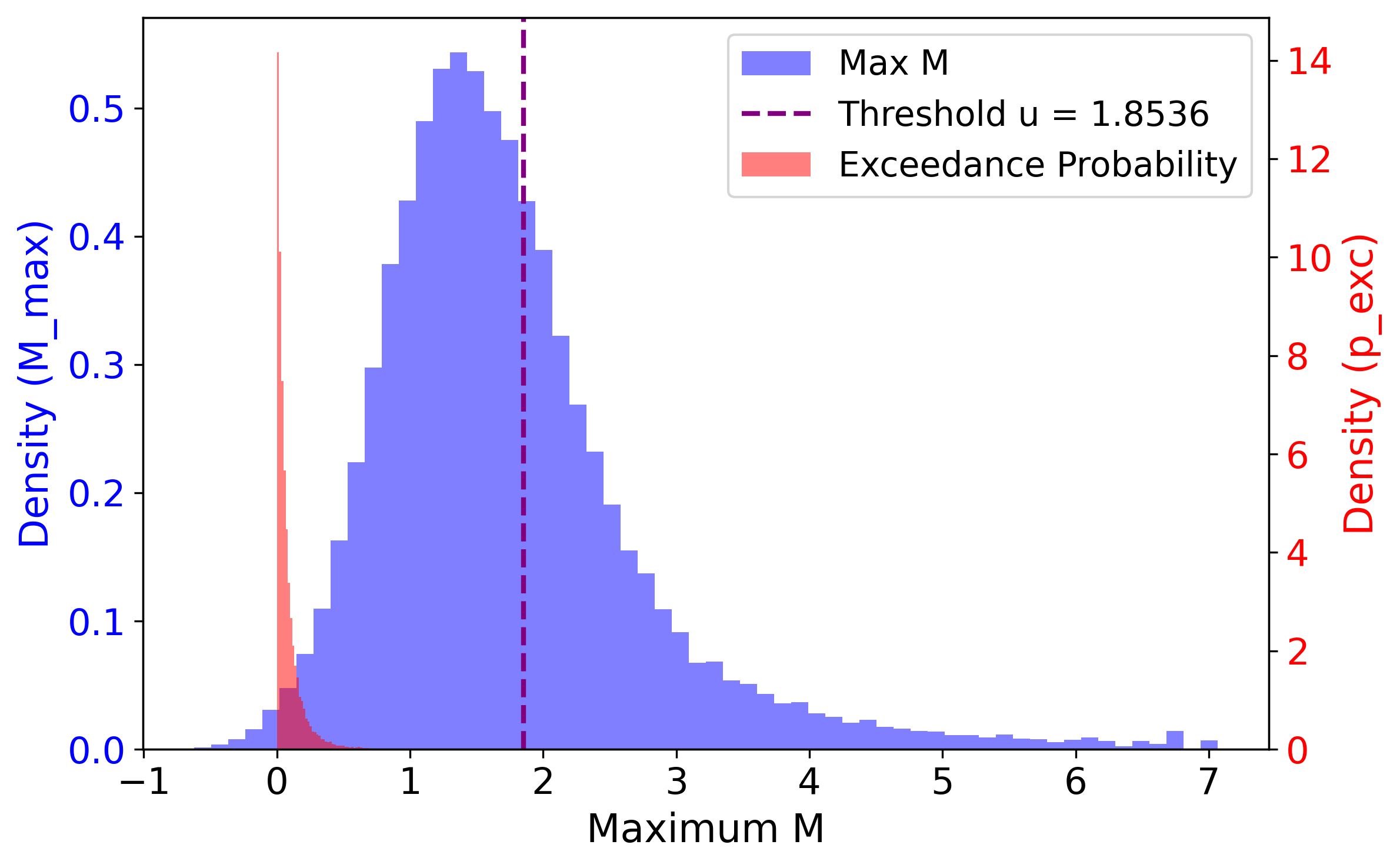}
\caption{Distribution of maximum interaction values ($M_{\text{max}}$). Dashed line: 97th percentile ($u=1.8536$).}
\label{fig:maxM}
\end{figure}

We evaluated AV behavior using the maximum interaction value ($M_{\text{max}}$) at each timestep, representing the most critical agent–AV proximity in multi-agent contexts. This metric captures the highest momentary collision risk. Figure~\ref{fig:maxM} shows that most operations occur in safe zones, though a notable fraction of timesteps exceed the 97th percentile ($u=1.85$), indicating near-conflict conditions.

\subsubsection{Headway}
\begin{figure}[ht!]
\centering
\includegraphics[width=0.56\linewidth]{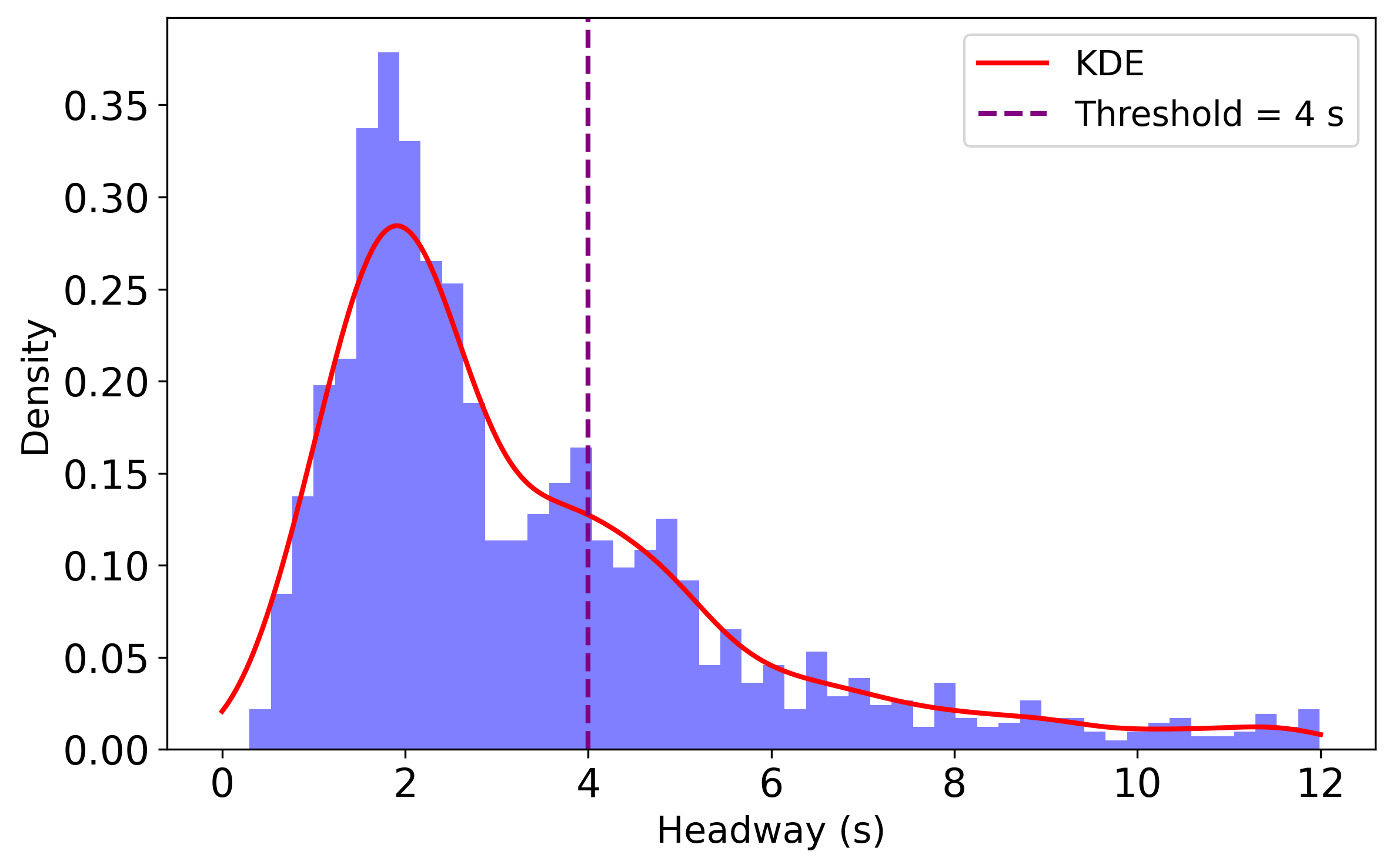}
\caption{Distribution of time headways across all AVs. Dashed line: 4~s (relaxed following).}
\label{fig:headway_dist}
\end{figure}

Figure~\ref{fig:headway_dist} shows time headways concentrated between 1–3~s, with longer headways ($>4$~s) mostly in Foggy Bottom due to lower speeds and complex urban conditions. These patterns indicate conservative AV behavior.

\subsubsection{String Stability}
\begin{figure}[ht!]
\centering
\includegraphics[width=0.56\linewidth]{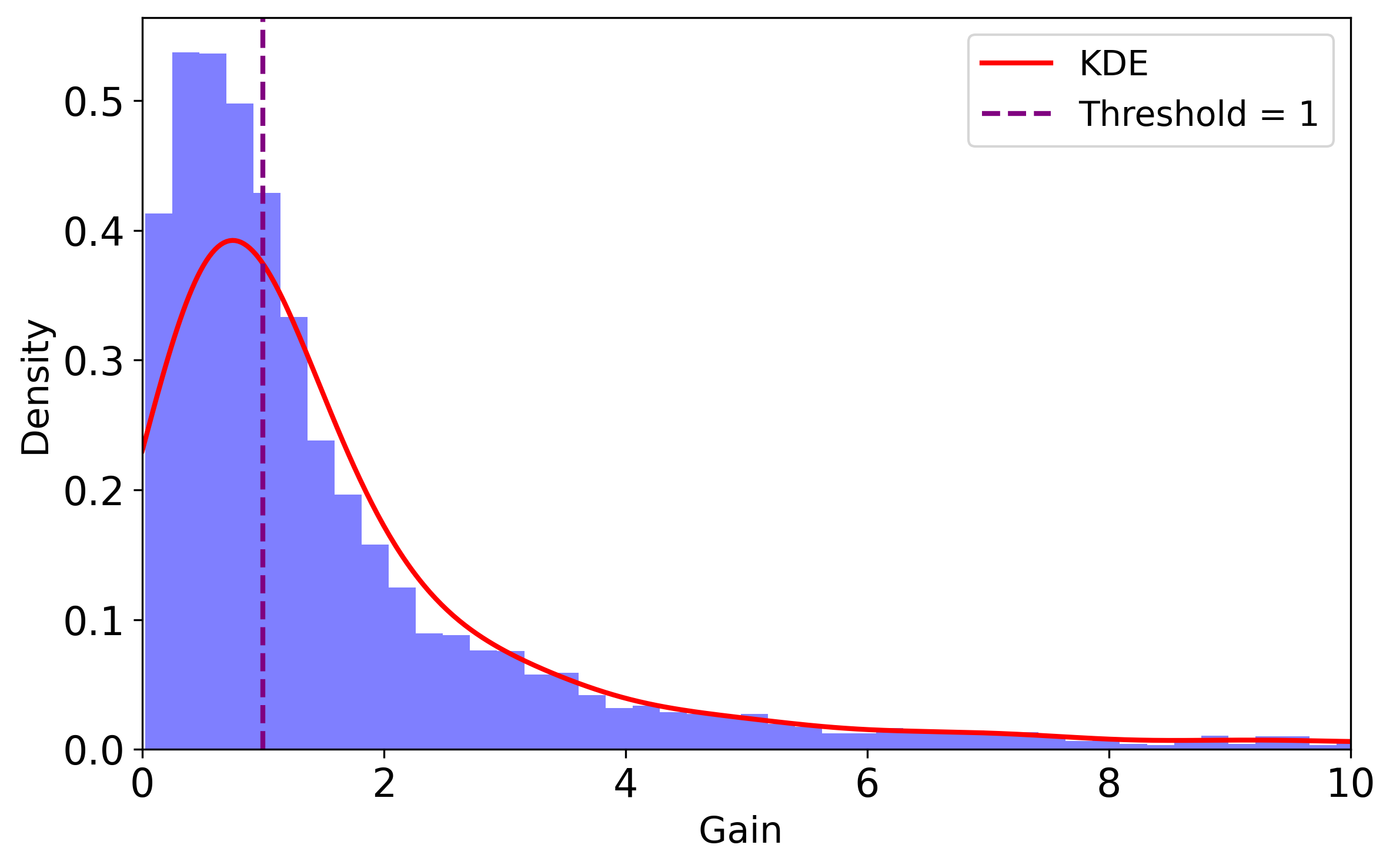}
\caption{Distribution of first follower string stability gain. Dashed line: Gain = 1.}
\label{fig:gain_dist}
\end{figure}

As shown in Figure~\ref{fig:gain_dist}, follower gain values cluster below one, meaning AVs generally dampen acceleration disturbances. The long upper tail ($>1$) reveals occasional amplification, suggesting opportunities for refining longitudinal control for greater stability.

\subsubsection{Jerk}
\begin{figure}[ht!]
\centering
\includegraphics[width=0.56\linewidth]{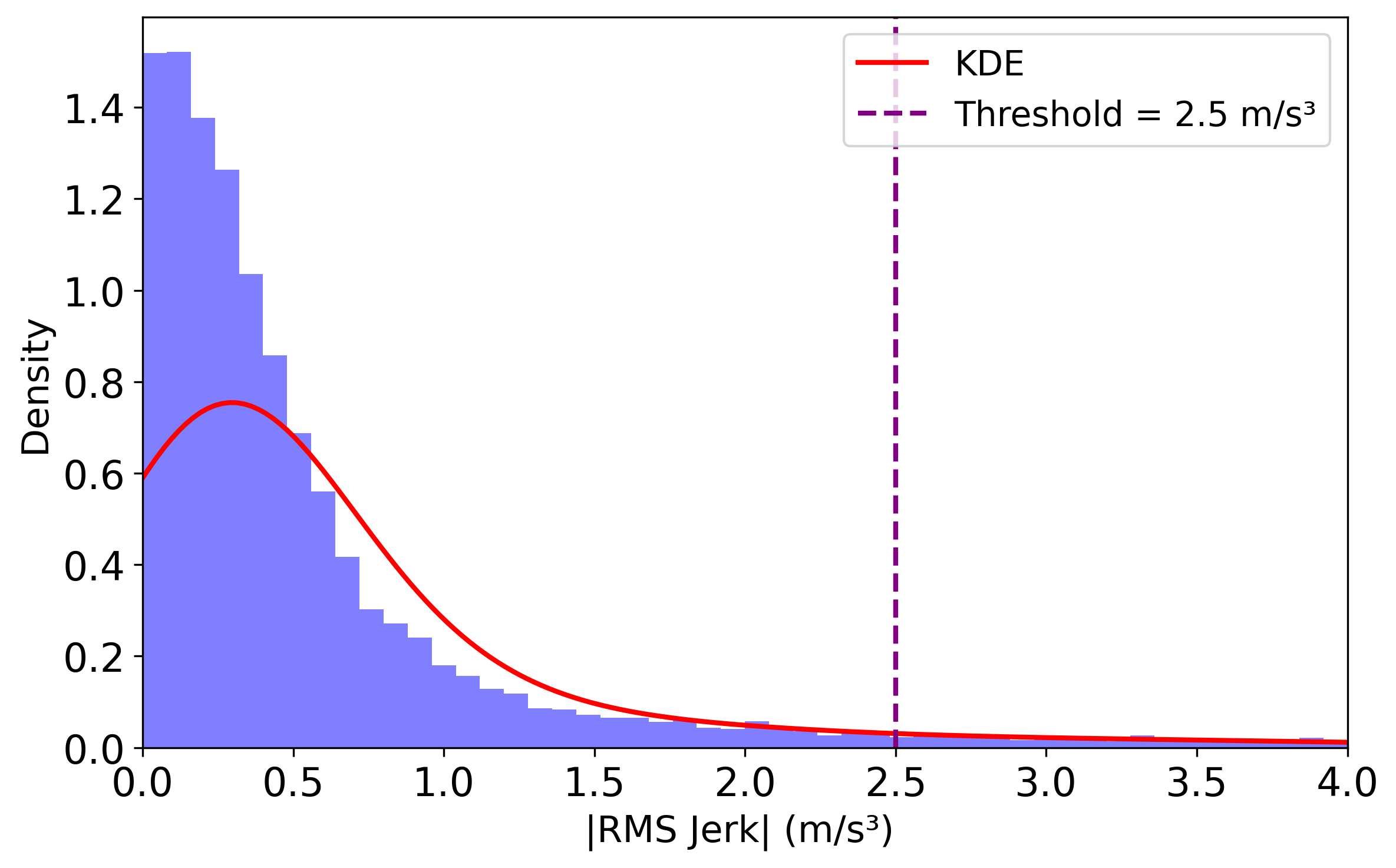}
\caption{Distribution of jerk. Dashed line: $2.5~\mathrm{m/s^3}$.}
\label{fig:jerk_dist}
\end{figure}

Most AVs exhibit low jerk (Figure~\ref{fig:jerk_dist}), however, $\sim$75\% of AVs exceed $2.5\,\mathrm{m/s^3}$ at some point. This suggests room for improvement in efficiency and interaction.

\subsubsection{Deceleration Intensity}
\begin{figure}[ht!]
\centering
\includegraphics[width=0.56\linewidth]{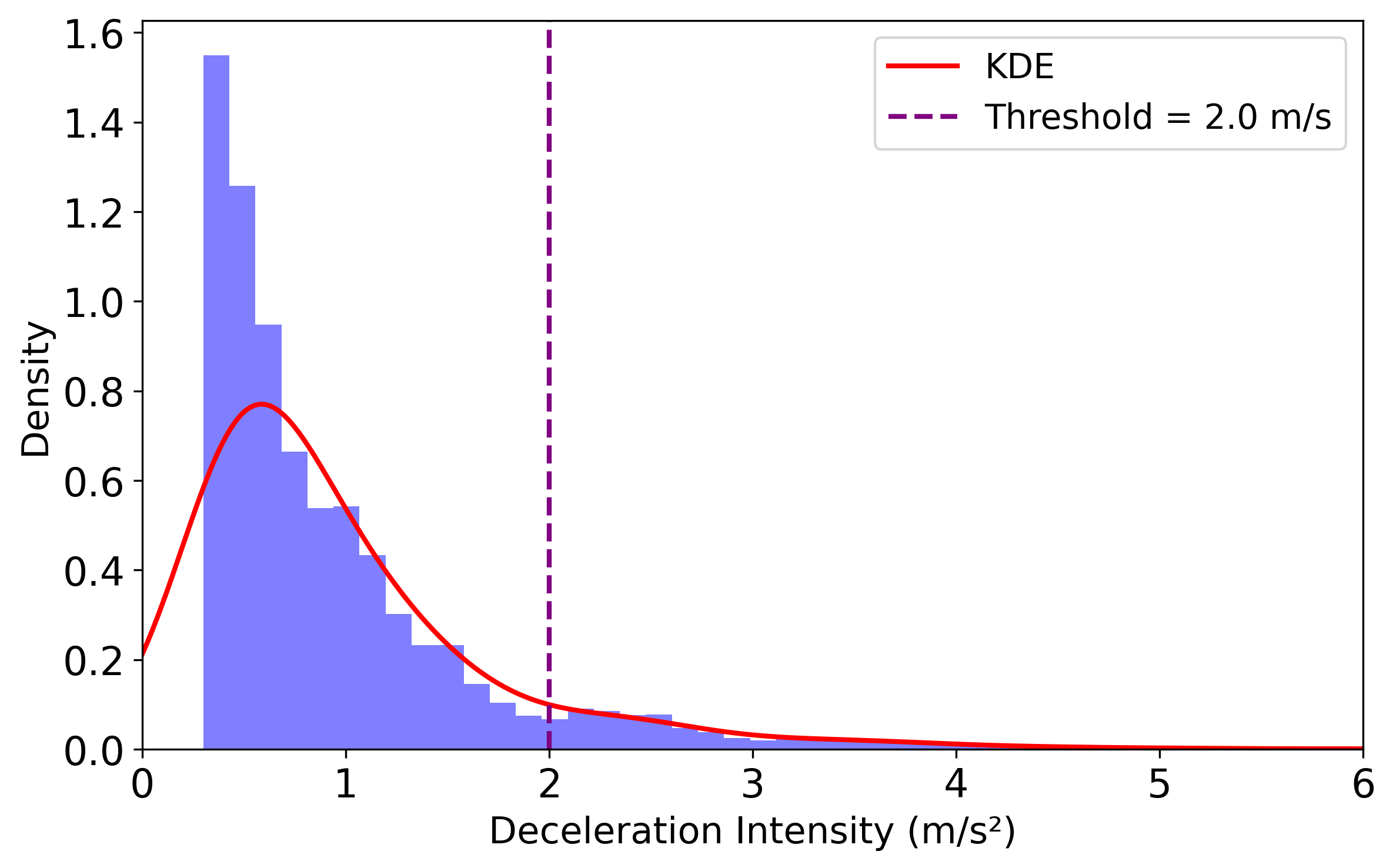}
\caption{Distribution of deceleration. Dashed line: $2.0~\mathrm{m/s^2}$.}
\label{fig:decel_dist}
\end{figure}

Deceleration magnitudes (Figure~\ref{fig:decel_dist}) peak below $1~\mathrm{m/s^2}$, reflecting anticipatory braking. Rare high-intensity events ($>2.0~\mathrm{m/s^2}$) occur across all AVs, representing short emergency responses that merit further control tuning.

\subsection{Empirical Pareto Optimality Analysis}

Behavioral metrics were grouped into three dimensions: safety, efficiency, and interaction. Composite scores for each dimension were computed as described previously, and Pareto dominance was applied to jointly maximize $(S,E,I)$.

Only 43 timesteps ($0.23\%$) were Pareto-optimal, confirming that optimization across all objectives is rare in naturalistic AV behavior. Table~\ref{tab:pareto_summary} summarizes the mean composite scores.

\begin{table}[ht!]
\centering
\caption{Mean composite scores by Pareto status.}
\label{tab:pareto_summary}
\begin{tabular*}{\linewidth}{@{\extracolsep{\fill}}lccc}
\toprule
Pareto Status & Safety & Efficiency & Interaction \\
\midrule
Non-Pareto & 0.760 & 0.798 & 0.502 \\
Pareto-optimal & 0.920 & 0.944 & 0.756 \\
\bottomrule
\end{tabular*}
\end{table}

\begin{figure}[ht!]
    \centering
    \includegraphics[width=0.95\linewidth]{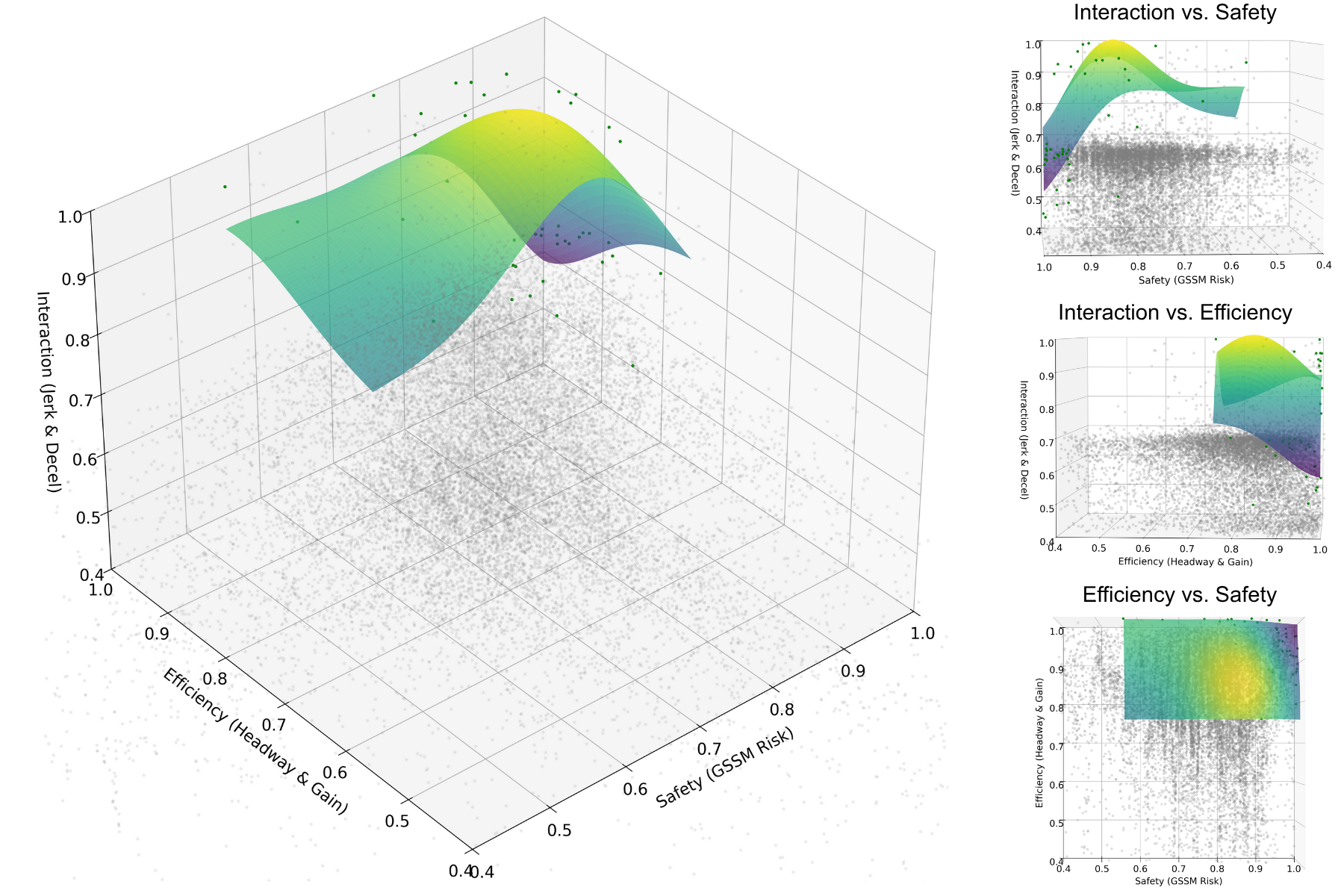}
    \caption{Empirical Pareto frontier in $(S,E,I)$ space. Green points denote Pareto-optimal states; grey points are dominated observations; the surface shows the GPR approximation.}
    \label{fig:pareto}
\end{figure}

The empirical Pareto frontier (Figure~\ref{fig:pareto}) forms a smooth surface toward the ideal $(1,1,1)$ point, representing the attainable trade-off boundary among the three objectives.

We quantified \textit{headroom}: deviation from the Pareto frontier, for each objective. Median headroom was highest for interaction ($0.45$), followed by safety ($0.19$) and efficiency ($0.15$).  

From a control perspective, these results provide a data-driven basis for adaptive multi-objective tuning: AVs could iteratively minimize headroom to move their operating point toward the empirical Pareto frontier, balancing safety, efficiency, and interaction in real time.

\section{Discussion and Implications}

The empirical Pareto surface represents the attainable balance among safety, efficiency, and interaction within naturalistic AV behavior. Unlike discrete or isolated optimal states, this surface defines a continuous trade-off manifold that embodies how improvements in one objective constrain others. It provides a geometric representation of multi-objective equilibrium rather than a single ``best'' policy.

\begin{figure}[ht!]
\centering
\includegraphics[width=0.65\linewidth]{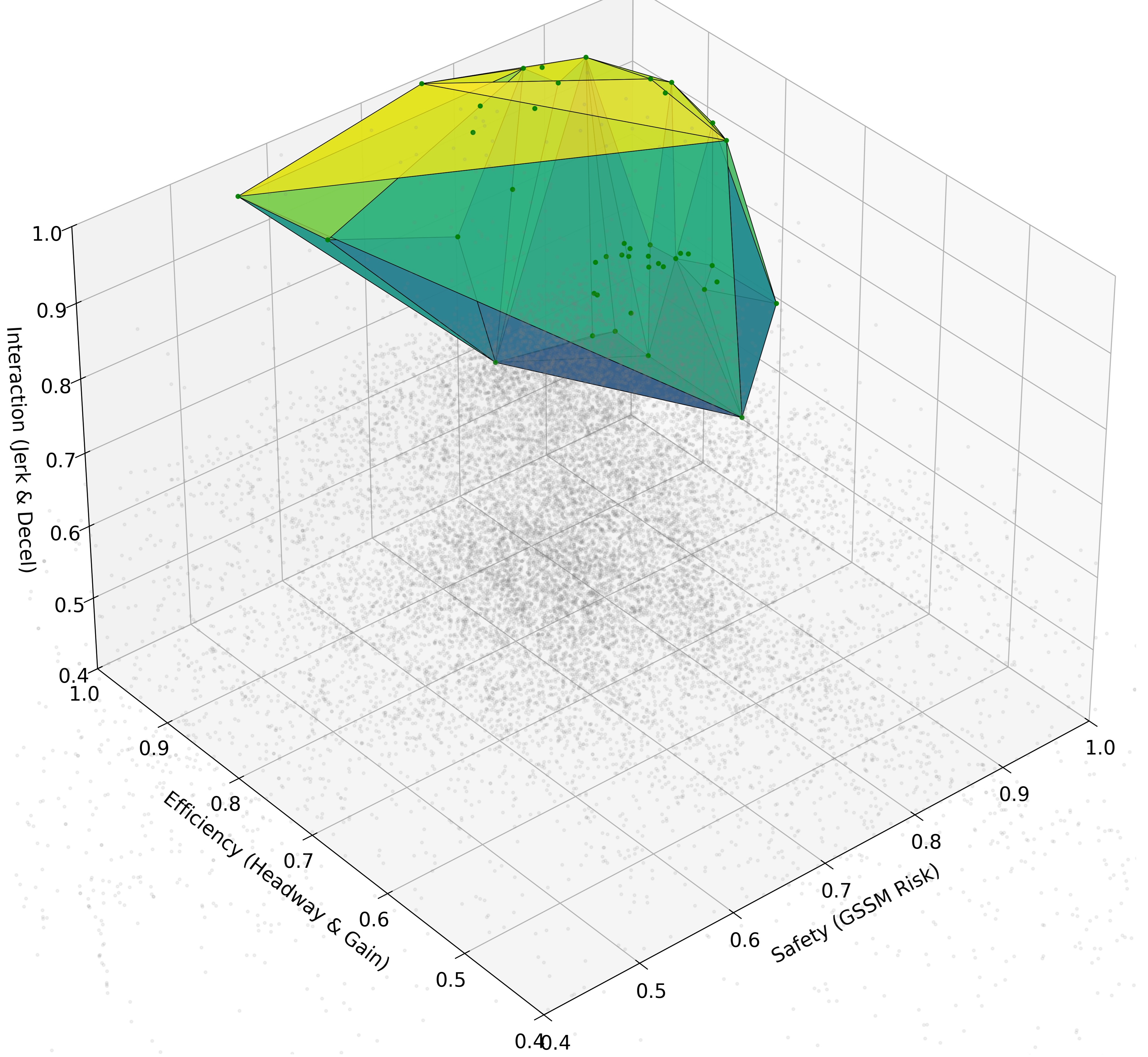}
\caption{Convex hull representation of the empirical Pareto frontier, showing the sharp, piecewise structure used for comparison with the smoothed GPR surface.}
\label{fig:pareto_convexhull}
\end{figure}

Compared to a convex hull (Figure~\ref{fig:pareto_convexhull}), the GPR surface offers smoother continuity and differentiability. A convex hull forms jagged, non-differentiable faces that exaggerate noise or underrepresent optimality near sharp edges. The GPR surface, learned with a composite Radial Basis Function (RBF) kernel, attenuates noise-induced extremes while preserving high-fidelity structure. Smoothing eliminates discontinuities that could yield ambiguous nearest-surface projections when a point lies near two adjacent hull faces. Overshoot tests confirmed numerical stability (max 0.0124, mean 0.0061; 3.4\% of grid exceeding 1). These were mitigated through kernel tuning and normalization, ensuring a realistic surface.

The framework is minimally invasive in that it does not modify the underlying trajectories or impose behavioral priors. It is modular and reproducible: starting from basic kinematic and positional data, a researcher can reproduce every stage, including metric derivation, normalization, imputation, and learning surface construction. The process generalizes to any trajectory dataset or objective set, enabling researchers to define alternative metrics beyond the five analyzed here.

Potential applications include Pareto-informed RL and end-to-end autonomous driving, where learned surfaces can guide policy updates or adaptive control. Examples include Pareto-optimal actor--critic driving frameworks \cite{9564464}, evolutionary adversarial trajectory optimization \cite{jiao2025evadrive}, Pareto set learning for multi-objective RL \cite{liu2025MORL}, and adaptive multi-objective driving personalization \cite{surmann2025}. These studies highlight the growing value of Pareto-guided learning, and the present work contributes by providing an empirical, data-driven foundation for constructing such multi-dimensional learning spaces from  trajectory data.

\section{Conclusion and Future Work}
This study introduced an empirical learning framework for mapping the multi-objective trade-offs among safety, efficiency, and interaction in naturalistic automated vehicle (AV) behavior. By deriving unified metrics from trajectory data and constructing an empirical Pareto surface, the analysis revealed that only a small fraction of AV states achieve near-optimal balance, with interaction showing the greatest headroom.

The work is limited by the rarity and scope of observed AVs. Foggy Bottom sequences include only left-turn maneuvers, while I-395 AVs operate mostly along a single lane.

Future efforts will integrate the learned Pareto surface into simulation and control contexts, using it as a reward structure or reference for end-to-end driving models. Expanding the framework to additional datasets and traffic contexts will further validate its generality and quantify performance gains from Pareto-informed decision making.

\bibliographystyle{IEEEtran}
\bibliography{references}

\end{document}